\documentclass{ieeeaccess}
\usepackage{cite}
\usepackage{amsmath,amssymb,amsfonts}
\usepackage{algorithm}
\usepackage{algorithmic}
\usepackage{graphicx}
\usepackage{textcomp}
\usepackage{booktabs}
\usepackage{array}

\usepackage{listings}
\usepackage{float}
\usepackage{microtype}
\usepackage{graphicx}
\usepackage{subcaption}
\usepackage{booktabs} 

\usepackage{xspace}
\usepackage{float}
\usepackage{stfloats}
\usepackage{wrapfig}
\usepackage[pagebackref=true,breaklinks=true,colorlinks,bookmarks=false]{hyperref}
\usepackage{multirow}
\usepackage{lipsum}
\usepackage{mathtools}
\usepackage{cuted}
\usepackage{multirow}
\usepackage{color}
\usepackage{colortbl}
\usepackage[frozencache,cachedir=.]{minted}
\definecolor{Green}{RGB}{102,252,102}
\definecolor{OrangeRed}{rgb}{1.0, 0.27, 0.0}

\definecolor{highlight}{rgb}{0.9, 0.9, 0.9}
\usepackage{xcolor}
\usepackage{soul}

\newif\ifhighlight
\highlightfalse
\sethlcolor{yellow!30}

\ifhighlight
    \newcommand{\rev}[1]{\hl{#1}}
\else
    \newcommand{\rev}[1]{#1}
\fi

\usepackage{mathtools,amssymb,lipsum}
\newcommand{\PreserveBackslash}[1]{\let\temp=\\#1\let\\=\temp}
\newcolumntype{C}[1]{>{\centering\arraybackslash}p{#1}}

\def\BibTeX{{\rm B\kern-.05em{\sc i\kern-.025em b}\kern-.08em
    T\kern-.1667em\lower.7ex\hbox{E}\kern-.125emX}}

\begin{document}
\history{Received 6 June 2026, accepted 22 July 2026, date of publication 12 August 2026, date of current version 19 August 2026.}
\doi{0.1109/ACCESS.2026.372302}

\title{Metric-Driven Reinforcement Learning for Object Hallucination Reduction in LVLM-Based Image Captioning}

\author{\uppercase{Thanh Nguyen},  \uppercase{Hee Suk Yoon}, \uppercase{Tung Minh Luu}, \uppercase{Eunseop Yoon}, \uppercase{Chang D. Yoo}}

\address[1]{School of Electrical Engineering, Korea Advanced Institute of Science and Technology, Daejeon 34141, Republic of Korea}


\markboth
{Author \headeretal: Preparation of Papers for IEEE TRANSACTIONS and JOURNALS}
{Author \headeretal: Preparation of Papers for IEEE TRANSACTIONS and JOURNALS}

\corresp{Corresponding author: Chang D. Yoo (e-mail: cd\_yoo@kaist.ac.kr).}

\begin{abstract}
Large Vision-Language Models (LVLMs) have excelled in joint visual and language understanding, particularly in image captioning. However, LVLMs remain prone to object hallucination, describing non-existent entities in the given image. Previous remedies, including dataset-enhanced supervised training and reinforcement learning from human feedback, show promise but rely heavily on costly human involvement, limiting scalability.
To address this, we propose a reinforcement learning framework to mitigate object hallucination in LVLM-based image captioning, driven entirely by natural language processing (NLP) metrics. While optimizing reinforcement learning solely on NLP metrics offers scalability, it also introduces pitfalls: reward hacking, sparse feedback, and costly training. Our framework sidesteps these by introducing an Object Consistency Score (OCS) capturing both correctness and coverage of objects, incorporating KL-regularization for denser token-level guidance, and designing a lightweight PPO variant that fine-tunes adapters atop a frozen LVLM backbone, achieving efficiency and stability. Extensive experiments on the InstructBLIP baseline demonstrate the effectiveness and generalizability of our framework, achieving a substantial hallucination reduction of up to 41\% on the COCO dataset and 29\% on the Visual Genome. Furthermore, the framework can enhance caption quality by incorporating complementary evaluation metrics such as BERTScore and METEOR. These results reveal a broader insight in multimodal learning: LVLMs can effectively self-correct hallucinations using metric feedback, motivating deeper exploration into scalable reinforcement frameworks that significantly reduce human effort.   

\end{abstract}


\titlepgskip=-15pt

\maketitle

\section{Introduction}

\begin{figure*}[htp]
    \centering
    \begin{subfigure}[b]{0.2\linewidth}
        \centering
        \includegraphics[width=\linewidth,height=5.5cm]{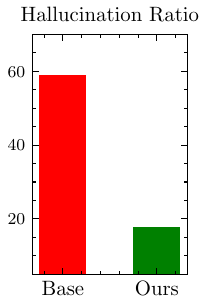}
        \caption{Sentence hallucination ratio measured in $\text{CHAIR}_s$}
    \end{subfigure}
    \hspace{0.05\linewidth} 
    \begin{subfigure}[b]{0.7\linewidth}
        \centering
        \includegraphics[width=\linewidth,height=5.5cm]{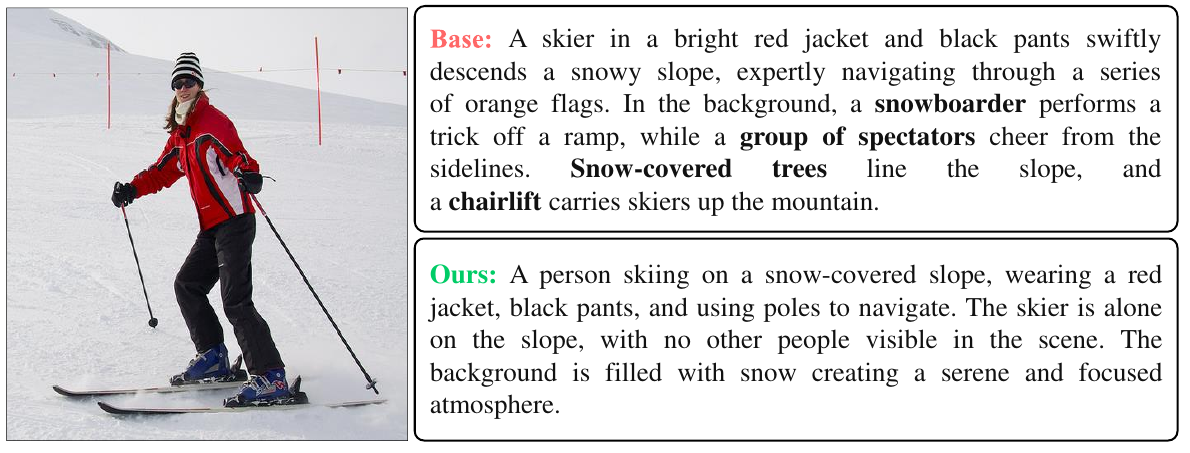}
        \caption{A detailed caption example from COCO dataset for baseline (\textcolor{OrangeRed}{Base}) vs fine-tuned (\textcolor{Green}{Ours}). Bold objects are hallucinated ones by LVLMs.}
    \end{subfigure}
    \caption{Quantitative and qualitative comparison between \textcolor{OrangeRed}{Base} (InstructBLIP) and \textcolor{Green}{Ours}: Chart (a) shows a significant 41\% reduction in object hallucination on the COCO dataset using \textcolor{Green}{Ours}. Figure (b) presents an example where the \textcolor{OrangeRed}{Base} model produces a caption with substantial object hallucination, while the \textcolor{Green}{Ours} model provides an accurate description without hallucinated objects.}
    \label{fig:DefensePerformance}
\end{figure*}

Large Vision-Language Models (LVLMs) have become increasingly prominent due to their ability to perform joint visual and language understanding tasks \cite{li2025survey}. 
Among these, image captioning has emerged as a key application where LVLMs consistently outperform smaller models by generating highly detailed and contextually rich captions \cite{dai2305instructblip,zhu2023minigpt,li2023blip}. Despite these advancements, LVLMs still struggle with a critical challenge: object hallucination \cite{rohrbach2018object,biten2022let}. This occurs when captions include references to objects that do not exist in the corresponding image, particularly in longer, more detailed descriptions, as shown in Figure \ref{fig:DefensePerformance}. Object hallucination not only undermines the credibility of these models but also hinders their broader application in fields that require high precision, such as autonomous systems \cite{yaqoob2019autonomous} and medical imaging \cite{chambon2022adapting}. 

Addressing object hallucination has been a major focus in recent research efforts \cite{liu2024survey,zhou2023analyzing, li2023evaluating, dai2022plausible, favero2024multi,liu2024paying,huang2024opera}. Among existing solutions, training-based approaches remain the high effective paradigm for mitigating hallucination in LVLMs and can be broadly divided into two categories: (1) improving dataset quality and fine-tuning LVLMs \cite{gunjal2023detecting, liu2023aligning} and (2) leveraging Reinforcement Learning from Human Feedback (RLHF) to reduce hallucinations \cite{sun2023aligning,gunjal2024detecting,yu2024rlhf}. Despite their potential, these methods still face significant challenges, as gathering large volumes of high-quality examples \cite{gunjal2024detecting,you2023ferret,zhang2024recognize} or obtaining accurate human feedback for RLHF fine-tuning remains a time-consuming and labor-intensive process that requires considerable human expertise and effort.

To address these limitations, we propose a metric-driven reinforcement learning framework for mitigating object hallucination in LVLM-based image captioning. The central idea is to employ hallucination-sensitive metrics (e.g., CHAIR\cite{rohrbach2018chair}) to automatically evaluate hallucinations and fine-tune the model accordingly. Since these metrics are often non-differentiable, direct optimization via supervised learning is infeasible. To overcome this, we formulate captioning as a Markov Decision Process (MDP) and leverage reinforcement learning, where non-differentiable metrics are incorporated as rewards.

However, fine-tuning under this setting introduces three key pitfalls:

(1) Reward hacking: Conventional hallucination metrics (e.g., CHAIR) often primarily count incorrectly generated objects, which LVLMs can exploit by producing captions with fewer objects. To address this, we propose a metric named \textit{Object Consistency Score} (OCS), which jointly considers correctness and coverage of objects, providing a more balanced reward signal.

(2) Sparse feedback: Most NLP metrics are computed only after the full caption is generated, resulting in sparse rewards and making RL optimization difficult, especially with long captions. To overcome this, we incorporate a KL divergence penalty that encourages generated tokens to remain close to the pretrained model token distribution. This approach stabilizes training by providing denser per-token feedback throughout the generation process and helps preserve caption quality by maintaining alignment with the pretrained model's prior.

(3) High computational and memory cost of fine-tuning LVLMs: Standard PPO-based fine-tuning requires maintaining a policy, a value function, and a reference model—each comparable in size to the LVLM—making training extremely resource-intensive. To reduce overhead, we propose a compact PPO variant where the policy, value function, and reference model share the same frozen backbone language model. Only lightweight adapters are trained, ensuring efficiency and compatibility with recent LVLM fine-tuning techniques such as prompt tuning.

 Extensive experiments show that our method reduces hallucination by up to 41\% compared to the InstructBLIP baseline, while preserving object coverage and caption quality on the COCO dataset. Moreover, the framework generalizes well to more complex datasets—achieving up to 29\% hallucination reduction on the Visual Genome—and can seamlessly incorporate additional NLP metrics such as BERTScore and METEOR for further quality improvement. Importantly, it achieves comparable or superior performance to prior methods without human feedback, offering a scalable and efficient solution for enhancing LVLMs in image captioning. 

\rev{Overall, the main contributions of this work are summarized as follows:}

   \begin{itemize}
    \item \textbf{Metric-driven RL framework:} We propose a reinforcement learning framework for mitigating object hallucination in LVLM-based image captioning, leveraging fully automatic NLP metrics and eliminating the need for human feedback.

    \item \textbf{KL-regularized reward with Object Consistency Score (OCS):} We design a reward function that combines token-level KL regularization with a novel Object Consistency Score (OCS). OCS is an F1-based metric that balances object precision and recall, encouraging both correctness and coverage while mitigating reward hacking observed in hallucination-only objectives.
    
   \item \textbf{Efficient PPO with shared backbone:} We develop a parameter-efficient PPO-based fine-tuning strategy in which the policy, value, and reference networks share a frozen LVLM backbone, improving training stability and efficiency.
    \item \textbf{Comprehensive empirical validation:} We conduct extensive experiments on COCO and Visual Genome, demonstrating significant reductions in hallucination and strong generalization across different LVLM architectures.
\end{itemize}

Our results reveal an important perspective in multimodal learning: metric-guided feedback empowers LVLMs to self-correct hallucinations without human supervision, motivating future exploration of reinforcement frameworks that achieve scalability with minimal human involvement.

The remaining sections are organized as follows: Section 2 reviews related work. Section 3 details our proposed methodology, including the formulation of the Markov Decision Process (MDP) for image captioning, the design of the reward function, and the modeling of reinforcement learning networks, followed by fine-tuning the model through solving the MDP with RL. Section 4 presents the experimental results, and Section 5 reports the ablation studies. Section 6 provides discussions, while Section 7 concludes the paper and outlines directions for future research.

\section{Related Work}

\textbf{Large Vision Language Model:} The rapid advancements in Large Language Models (LLMs) \cite{zhao2023survey,chung2024scaling}, combined with advances in computer vision, have spurred the development of large vision-language models (LVLMs) \cite{li2025survey,liu2023visual,zhu2023minigpt,sun2023aligning,ye2023mplug}. LVLMs integrate an LLM with a pretrained visual encoder to form end-to-end models capable of generating contextually relevant text from visual stimuli \cite{zhang2024vision}.  
Typical LVLM architectures consist of three components: a visual encoder, a modality connection module, and an LLM. 
The visual encoder, often adapted from CLIP \cite{radford2021learning,li2022blip,li2023blip}, transforms images into visual tokens. These tokens are then aligned with the LLM’s word embedding space through a connection module, which may employ cross attention \cite{alayrac2022flamingo}, simple projection layers (linear or multilayer perceptron) \cite{liu2023visual,yu2023mm}, or Q-former modules \cite{li2023blip2,dai2305instructblip}. Finally, the LLM then acts as the central reasoning unit, integrating aligned visual and textual information to generate responses. Training is commonly performed in two stages \cite{ye2023mplug}: first optimizing the vision encoder and connection module, followed by fine-tuning the LLM with lightweight adaptation techniques such as LoRA \cite{hu2022lora}. 
Notably, Q-former-based LVLMs can achieve strong performance without requiring LLM fine-tuning \cite{li2023blip2,dai2305instructblip}.  

In this work, we adopt Q-former-based InstructBLIP \cite{dai2305instructblip} as our baseline to evaluate the effectiveness of our method for mitigating object hallucination in LVLMs.

\textbf{Object Hallucination in Vision Language Models:} Object hallucination refers to the generation of descriptions containing objects absent from the visual modality (i.e., the given image) \cite{rohrbach2018object}. Prior work has investigated this through evaluation and detection techniques \cite{petryk2024aloha,li2023evaluating,dai2022plausible,jing2023faithscore} and benchmark development \cite{ben2024mitigating,wang2023evaluation}, which confirm hallucination as a critical problem requiring mitigation. Early attempts to mitigate hallucination employed contrastive learning \cite{zeng2021multi,leng2024mitigating}, data augmentation \cite{kim2023exposing}, or non-training strategies \cite{favero2024multi,liu2024paying,huang2024opera}. However, these methods are less effective for LVLMs, particularly when there is a shift in data distribution, and often serve as lightweight complements to training-based approaches \cite{zhang2023language, zhou2023analyzing, liu2024survey}. Training-based approaches remain highly effective for LVLMs and can be broadly divided into two categories. The first focuses on improving dataset quality and fine-tuning models. 
High-quality datasets have been shown to reduce hallucination by providing more reliable supervision \cite{gunjal2023detecting,you2023ferret}, after which LVLMs are typically fine-tuned using supervised learning techniques \cite{zhou2023analyzing,zhai2023halle}. 
While effective, this direction heavily depends on the availability of curated datasets, which are expensive to construct and often limited in scope, thereby restricting scalability.  The second line of work employs Reinforcement Learning from Human Feedback (RLHF) \cite{sun2023aligning,stiennon2020learning,gunjal2024detecting,yu2024rlhf}, where human-annotated preferences are used to train reward models and guide LVLMs toward producing more faithful outputs. 
Although RLHF has demonstrated strong improvements, it requires large-scale human annotations, extensive preference comparisons, and costly iterative optimization pipelines. 
Consequently, both directions, despite their success, incur significant time and labor costs, limiting their practicality for large-scale or rapidly evolving applications.  

In contrast, we propose a conceptually distinct approach grounded in reinforcement learning with automatic metrics. By replacing human supervision with reliable automatic signals, our method offers an efficient and scalable solution that mitigates hallucination while remaining compatible with diverse LVLM architectures—eliminating the need for costly human feedback.

\textbf{Reinforcement Learning for NLP:} Reinforcement Learning (RL) is a well-established framework with strong success in domains such as robotics and control \cite{finn2017model,nguyen2021robust,nguyenone,nguyen2026fast,luu2024mitigating,luu2022utilizing,nguyen2024towards,nguyen2024uncertainty}. More recently, it has become an increasingly important technique for improving language models, particularly in Natural Language Processing (NLP) tasks. RL has been successfully applied to dialogue \cite{li2016deep,yi2019towards,jaques2020human}, machine translation \cite{nguyen2017reinforcement,kiegeland2021revisiting,kreutzer2018can}, image captioning \cite{rennie2017self,ren2017deep}, summarization \cite{stiennon2020learning,wu2018learning,bohm2019better}, text-based games \cite{narasimhan2015language,hausknecht2020interactive}, and vision understanding \cite{das2017learning,nguyen2021sample,luu2022visual,nguyen2023dimcl}. In NLP training paradigm, models are optimized through an RL algorithm, wherein the reward signal is derived from either human feedback \cite{kreutzer2018can,jaques2020human,stiennon2020learning} or NLP evaluation metrics, such as ROUGE for summarization \cite{wu2018learning} or BLUE for translation \cite{nguyen2017reinforcement,kiegeland2021revisiting}. These reward mechanisms enable the models to iteratively improve and fine-tune their performance based on the quality of generated outputs. While RL has proven effective in Natural Language Processing (NLP), its exploration in Vision Large Language Models (LVLMs) for captioning is not well-established. Our work pushes the boundaries in this direction by leveraging RL to address the challenge of object hallucination in LVLMs. We tackle intricate issues specific to this context, including high computational costs, sparse rewards, and extended temporal horizons.

\textbf{Fine-tuning LVLMs with Adapters:} Fine-tuning the entire model for LVLM demands extensive memory and computational resources. To address this challenge, various Parameter Efficient Fine-Tuning (PEFT) methods have emerged as cost-effective alternatives. These methods include  prompt tuning \cite{lester2021power,li2021prefix,qin2021learning}, tuning hidden states (IA3
) \cite{liu2022few}, employing Low-rank Adapters (LoRA) \cite{hu2022lora,dettmers2023qlora},  incorporating full layers  \cite{houlsby2019parameter}, tuning
biases \cite{zaken2021bitfit}, learning weight masks based on Fisher information \cite{sung2021training}, and leveraging combinations of these approaches \cite{karimi2021compacter}. In our study, we demonstrate the effectiveness of prompt tuning-the simplest yet effective approach-for the task at hand, leaving the exploration of other PEFT methods for future work.

\section{Methodology}

In this session, we will sequentially cover the following topics: (1) casting the image captioning within the framework of a Markov Decision Process (MDP); (2) designing the dedicated reward function for object hallucination reduction; (3) modeling RL networks; (4) fine-tuning the LVLM by solving the MDP through Proximal Policy Optimization (PPO).

\begin{figure*}[htb]
    \centering
    \includegraphics[width=0.8\textwidth]{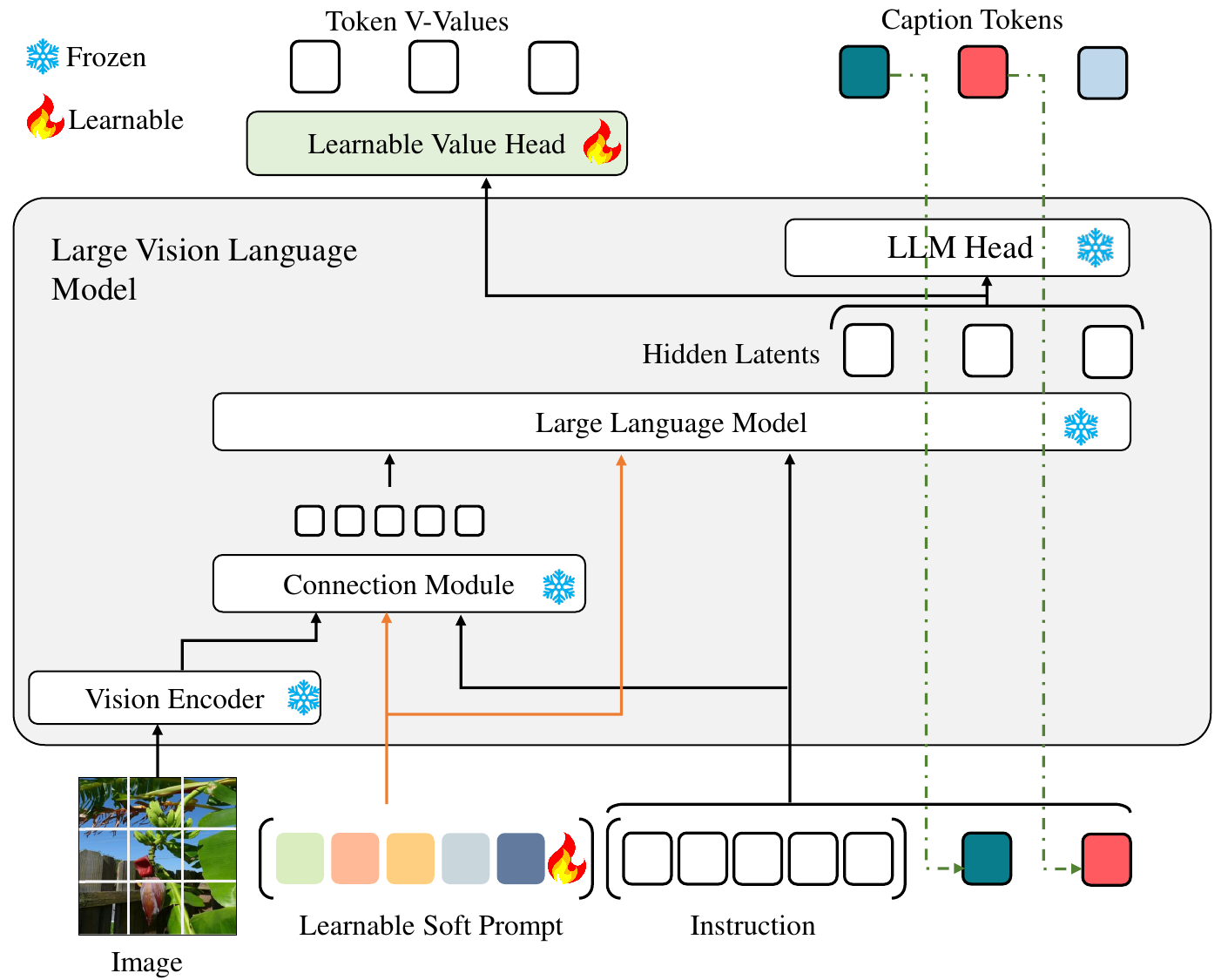}
    \caption{Detailed architecture of our framework. Specifically, the Policy Network is crafted by augmenting the shared LVLM with delicately learnable soft prompts. Meanwhile, the Value Network is formed by replacing the LLM Head with a Linear Value head. Notably, all parameters of the LVLM remain shared and frozen, with only a very small fraction (less than 0.01\% LVLM weight) of trainable parameters added to the LVLM for the meticulous modeling of the policy network and value network. }
    \label{fig:detail_architecture}
\end{figure*}


\subsection{Markov Decision Process (MDP) for image captioning} 

Image captioning with LVLMs involves generating a natural language description of an image, which requires both visual understanding and linguistic generation. Traditionally, the task is formulated as supervised learning, where the model is fine-tuned to predict captions aligned with human-annotated ground truths. However, incorporating feedback from Natural Language Processing (NLP) metrics—most of which are non-differentiable—poses challenges for optimization within the supervised learning paradigm. Because caption generation is inherently sequential, this task can be effectively framed as a Markov Decision Process (MDP), in which the model iteratively decides the next token based on the current state—defined by the image and preceding text tokens. The MDP allows for the optimization of non-differentiable reward functions, such as those based on NLP metrics for reducing object hallucinations, through reinforcement learning (RL).

Mathematically, we formulate the image captioning as an MDP denoted by a tuple $(\mathcal{S}, \mathcal{A}, R, P, \gamma, H)$, where 
$\mathcal{S}$ is the state space, 
$\mathcal{A}$ is the action space, 
$R$ is the reward function, $P$ is the transition function, 
$\gamma$ is the discount factor, and $H$ is the horizon. The detailed definitions of each component are as follows: Each episode (i.e., a complete caption generation) in this MDP begins by sampling a datapoint $(Z,X, C)$ from our dataset $\mathcal{D}=\{(Z_i,X_i, C_i)\}_{i=1}^N$, where  $Z \in \mathcal{Z}$ represents the image, $X$ represents the input text (a.k.a. instruction) for LVLMs, and $C$  is the ground truth caption, which can be set to \textit{None} if no ground truth caption is available. Here, $X \in \mathcal{W}$ and $C \in \mathcal{W}$, where $\mathcal{W}$ denotes the set of all possible textual sequences formed from the vocabulary $\mathcal{V}$. The initial state $S_0= \left(Z,x_0, \cdots, x_m\right)$ consists the image $Z$ and the instruction $X=\left(x_0, \cdots, x_m\right)$, where $S_0 \in \mathcal{S}$ and the state space 
 $\mathcal{S} = \mathcal{Z} \times \mathcal{W}$  represents the set of all possible image–text concatenations. At each time step $t$, an action $a_t \in \mathcal{A}$, which corresponds to a token from our vocabulary $\mathcal{V}$, is sample from a policy $\pi(a_t|S_t)$ (i.e., the LVLM) where $S_{t}=\left(Z,x_0, \cdots, x_m, a_0, \cdots, a_{t-1}\right)$. Then, the transition function $P: \mathcal{S} \times \mathcal{A} \rightarrow \mathcal{S}$ deterministically appends the action $a_t$ to the end of the state $S_{t}$ to form the state $S_{t+1}$. This process continues until the end of the episode time step $T$ $(0 < T < H)$, either when the current time step $t$ reach the horizon or when an end-of-sentence (EOS) token is generated, resulting in a final state $S_{T+1}=\left(Z,x_0, \cdots, x_m, a_0, \cdots, a_{T}\right)$. At every step, a reward signal is emitted, calculated from the reward function $R$. This reward function is designed to address the objective of the task, such as reducing object hallucination in generated captions. The resulting MDP can be optimized through RL, where the goal is to identify the optimal LVLM policy $\pi^*$ that maximizes the expected cumulative reward, formulated as:

\begin{equation}
 \max_{\theta} \mathbb{E}_{(a_0, \dots, a_T) \sim \pi_\theta; S_0, C \sim \mathcal{D}} \left[ \sum_{t=0}^{T} \gamma^t R(S_t, a_t, C) \right]
\end{equation}
where $\gamma$ denotes the discount factor (e.g., 0.99) and $(a_0...a_T)$ (denoted as $\hat{C}$) is the generated caption  from the LVLM.

\subsection{Designing The Reward function}

To address hallucination within the MDP framework, the primary challenge is to design a reward function that effectively guides RL to optimize the LVLM for the desired goal. A straightforward approach is to incorporate hallucination-sensitive metrics, such as the Caption Hallucination Assessment with Image Relevance (CHAIR) \cite{rohrbach2018chair}, directly into the reward function.

CHAIR has become a widely recognized standard for quantifying object hallucination in image captioning tasks. This metric assesses the alignment between the objects depicted in the image and those referenced in the generated caption. Specifically, CHAIR evaluates the percentage of hallucinated objects—objects mentioned in the caption that do not appear in the image. There are two variants of \(\text{CHAIR}\) for measuring object hallucination: \( \text{CHAIR}_s \), which calculates the fraction of captions containing hallucinated objects out of the total number of captions, and \( \text{CHAIR}_i \), which measures the proportion of hallucinated objects in a single caption relative to the total number of objects in that caption. The former assesses hallucination at the caption level, while the latter operates at the object level. These metrics are mathematically defined as follows:
\begin{equation}
\text{CHAIR}_i = \frac{|\{ o_j \in \hat{O} \mid o_j \notin O \}|}{|\hat{O}|},
\end{equation}
where \( \hat{O} \) represents the set of objects in the generated caption and \( O \) represents the set of objects in the image. The term $ o_j \in \hat{O} \mid o_j \notin O$ refers to a hallucinated object (an object in the caption that is not in the image).
\begin{equation}
\text{CHAIR}_s = \frac{|\{ \hat{C}_k \mid \exists o_j \in \hat{O}_k \text{ s.t } o_j \notin O_k \}|}{|\hat{\mathcal{C}}|},
\end{equation}
where \( \hat{C}_k \) represents the \( k \)-th generated caption, \( \hat{O}_k \) is the set of objects in the \( k \)-th generated caption, and \( O_k \) is the set of objects in the corresponding image. \( \hat{\mathcal{C}} \) is the set of all generated captions.

It is clear from the formulation that $0 < \text{CHAIR}_i < 1$, with $\text{CHAIR}_i$ approaching 1 when all objects in the caption are hallucinated and approaching 0 when no hallucination occurs. To minimize hallucination, the reward function can be set as \( 1 - \text{CHAIR}_i \), which encourages reducing hallucinated objects. This formulation directly promotes increasing the precision of the generated objects relative to the total objects in the caption.

However, while this reward function is conceptually sound, the fine-tuned LVLM can exploit it by generating fewer objects in the caption, which might lead to a high reward despite missing relevant objects from the image. This strategy results in a reduction of object coverage and an incomplete caption, undermining the goal of capturing the full content of the image.

To mitigate this issue, we propose a metric named Object Consistency Score (OCS), a variant of the F1 score \cite{christen2023review}. The OCS offers a balanced evaluation of both precision and recall, ensuring that the reward function promotes comprehensive object coverage while maintaining the precision of the generated objects.

\begin{equation}
    \text{OCS} = \frac{2 \cdot \textit{Pre} \cdot \textit{Rec}}{\textit{Pre} + \textit{Rec}},
\end{equation}
where \textit{Pre} and \textit{Rec} are the precision and recall, respectively. The \textit{Pre} is defined as the ratio of correctly mentioned objects to all objects mentioned in the caption, which can be computed as:
\begin{equation}
\textit{Pre} = 1 - \text{CHAIR}_i = \frac{|\{ o_j \in \hat{O} \mid o_j \in O \}|}{|\hat{O}|},
\end{equation}
On the other hand, the \textit{Rec} measures the ratio of correctly mentioned objects to all objects in the image, calculated as:
\begin{equation}
\textit{Rec} = \frac{|\{ o_j \in \hat{O} \mid o_j \in O \}|}{|O|}.
\end{equation}
\rev{Recall that $\hat{O}$ represents the set of objects in the generated caption and $O$  represents the set of objects in the image.} The image object set can be obtained by applying an off-the-shelf object detection model (e.g., YOLOv8~\cite{varghese2024yolov8}) or by using metadata of the image provided in the dataset. Similarly, the caption object set can be extracted using handcrafted rules (as in the $\text{CHAIR}$ metric \cite{rohrbach2018chair}) or with GPT-assisted parsing (as employed in the CCEval metric \cite{zhai2023halle}).

\rev{In essence, the OCS measures how well the generated caption remains consistent with the visual objects present in the image, making it preferable to CHAIR. Specifically, CHAIR only penalizes objects that are incorrectly mentioned in the caption (i.e., objects that do not appear in the image). As a result, the model can achieve a high score by mentioning fewer objects, even if it misses many objects that are actually present. In contrast, OCS considers both (1) how many mentioned objects are correct and (2) how many ground-truth objects are covered, encouraging the model to mention all relevant objects while avoiding incorrect ones.}

The resulting reward function is: 
\begin{align}
    R \left(S_t,a_t,C\right) = 
    &\begin{cases} 
    &\text{OCS}(S_t,a_t) \text{  if } t = T \\
    &0  \text{ otherwise.} 
    \end{cases}
\end{align}

Optionally, in the setting where ground truth captions are available, two additional metrics can be integrated into the reward function to further enhance caption quality: $\text{METEOR}$ \cite{banerjee2005meteor} and $\text{BERTScore}$ \cite{zhang2019bertscore}. $\text{METEOR}$ evaluates the similarity between generated and reference texts (a.k.a. ground truth captions) based on n-grams and word order, ensuring structural and lexical alignment. Meanwhile, $\text{BERTScore}$ assesses semantic similarity using pretrained BERT embeddings, capturing underlying meaning accurately. Together, $\text{METEOR}$ and $\text{BERTScore}$ offer a comprehensive evaluation of caption quality, considering both surface-level and semantic aspects, thereby improving caption relevance to the ground truth.

The enhanced reward function is defined as:
\begin{align}
    &R \left(S_t,a_t, C\right) =\\ 
    &\begin{cases} 
    &\text{OCS}(S_t,a_t) + \alpha \text{METEOR}(S_t,a_t,C) \\ 
    & \quad + \beta \text{BERTScore}(S_t,a_t,C) \text{  if } t = T \\
    &0  \text{ otherwise.} 
    \end{cases}
\end{align}

\rev{Note that the balancing weight $\alpha$ for $\text{METEOR}$ should be relatively smaller than $\beta$  for $\text{BERTScore}$, as METEOR relies on n-gram matching and word order, which may bias the model toward shorter captions, particularly in datasets with short ground-truth references.}

It is noteworthy that the NLP metrics incorporated into the reward function are computed only after the full caption is generated, which results in sparse rewards. This sparsity makes RL optimization challenging, particularly for long captions. Furthermore, in settings where ground truth captions are unavailable, METEOR and BERTScore cannot be used, causing the LVLM to excessively focus on reducing hallucinations without considering the overall quality of the caption. As a result, the generated captions may become incoherent or grammatically incorrect, such as by producing a list of objects without proper context.

To address this issue, we incorporate a KL divergence penalty that constrains the generated token distribution to remain close to that of the pretrained model. This regularization provides denser feedback, stabilizes training, and preserves caption quality without relying on ground-truth captions. 

Formally, at each token generation step, we regularize the reward function with a token-level KL term, ensuring that the fine-tuned LVLM $\pi_\theta(a_t|S_t)$ does not deviate excessively from the pretrained reference model $\pi^r(a_t \mid S_t)$. This mechanism densifies the reward signal and maintains semantic alignment with the original model. The regularized reward is defined as:
\begin{equation}
\begin{aligned}
&\hat{R}\left(S_t, a_t, C\right) \\
&= R\left(S_t, a_t, C\right)
- \lambda\, \mathrm{KL}\!\left(\pi_\theta\left(a_t \mid S_t\right) \,\|\, \pi^r\left(a_t \mid S_t\right)\right),
\end{aligned}    
\end{equation}
where $\hat{R}$ denotes the KL-regularized reward, $\mathrm{KL}$ represents the Kullback–Leibler divergence, and $\lambda$ is the weighting coefficient controlling the regularization strength.

\subsection{Modeling Reinforcement Learning Networks}
\begin{figure}[htp]
    \centering
    \includegraphics[width=0.5\textwidth]{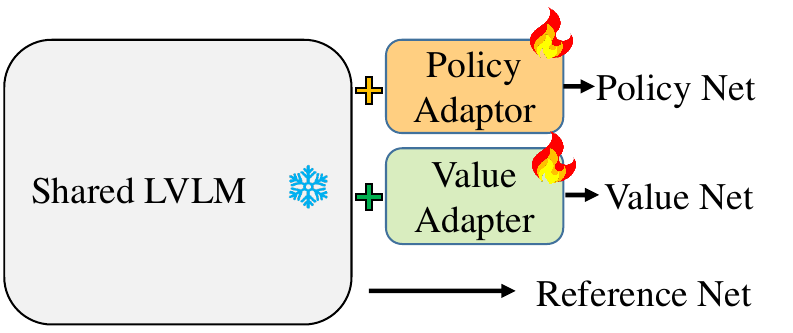}
    \caption{Simplified overview of the framework's network components: policy network, value network, and reference network. All networks share the same frozen LVLM as its foundation. The reference network mirrors the LVLM identically, while the value and policy networks incorporate a lightweight adapter into the shared LVLM.}
    \label{fig:simplifyNetworks}
\end{figure}

To fine-tune the LVLM with RL, we need a policy, a value network, and a reference network. Conventional approaches train these networks separately, each is comparable in size to the original LVLM, resulting in substantial computational overhead. To mitigate this, we design lightweight and efficient network variants. Figure \ref{fig:simplifyNetworks} displays the simplified overview of our network components. Specifically, each network builds upon the same frozen LVLM foundation. The reference network mirrors the LVLM identically, while the value and policy networks incorporate slender adapters alongside the LVLM. This approach optimizes computational resources and is compatible with various state-of-the-art Parameter-Efficient Fine-Tuning (PEFT) methods \cite{peft}, which rely on adapters.

For demonstration, we utilize \textit{Prompt Tuning} to assess the framework's effectiveness. Prompt Tuning offers an efficient and flexible method for controlling LVLM behavior by prefixing a learnable prompt to the instruction. By allowing the model to remain frozen while refining prompts, this approach reduces computational costs and provides task-specific adaptability \cite{lester2021power}. Mathematically, captioning is a sequence of token generation from the LVLM $\pi_\theta(a |S)$, with $\theta$ denoting the LVLM's weight. Prompt Tuning affects the LVLM's generation by providing additional context, which is achieved by prefixing a token sequence G to the instruction X, resulting in $\pi_\theta(a |G, S)$. Throughout, the model parameters $\theta$ remain unchanged. Optimal G selection can be achieved via manual exploration (\textit{Hard Prompting}) or by representing G with dedicated parameters $\phi$, refined through gradient descent (\textit{Soft Prompting}). This updates the conditional generation as $\pi_{\theta;\phi} (a |G, S)$, trainable by maximizing reward through backpropagation, with gradient updates solely applied to $\phi$, i.e., the learnable soft prompt.

 Figure \ref{fig:detail_architecture} illustrates the detailed architecture of the Augmented LVLM in our implementation. The Policy Network $\pi_{\theta;\phi} (a|G,S)$, is constructed by enhancing the shared LVLM with the delicately \textit{learnable soft prompt}. Concurrently, the Value Network $V_{\theta;\omega}(S)$ is created by substituting the LLM Head with a  Value Head, which is essentially a linear layer featuring a single output neuron. The reference network $\pi^r_\theta(a|S)$ remains identical to the original LVLM. Notably, all parameters of the LVLM persist as shared and frozen. Only an extremely small fraction (approximately 0.01\% LVLM weight) of trainable parameters is introduced to meticulously model the policy network and the value network.

\subsection{Fine-tuning model by solving the MDP}

Given the MDP and the RL networks, we fine-tune the augmented LVLM, i.e., the policy, using the Proximal Policy Optimization (PPO) algorithm \cite{schulman2017proximal}. Formally, this algorithm trains the policy $\pi_{\theta;\phi} (a|G,S)$ to maximize long-term discounted rewards over generated captions:
\begin{equation}
     \max_{\phi} \mathbb{E}_{a_t \sim \pi_{\theta;\phi }; S_0, C \sim \mathcal{D}}\left[\sum_{t=0}^T \gamma^t \hat{R}\left(S_t,a_t, C\right)\right].
\end{equation}
We define our V-value as follows:
\begin{equation}
 V^\pi(S_t)=\mathbb{E}_{a_t \sim \pi_{\theta;\phi }, S_0,C\sim \mathcal{D}}\left[\sum_{i=t}^T \gamma^i \hat{R}\left(S_i, a_i, C\right)\right] . 
\end{equation}

 We model this value function using the previously introduced value network $V_{\theta;\omega}$. With both the policy and value networks in place, the PPO algorithm \cite{schulman2017proximal} is then applied to fine-tune the policy parameters.

 \rev{Due to its compatibility with the standard PPO framework, training remains stable in practice. In particular, the KL divergence is effectively regulated around the target value, and no instability (e.g., reward collapse or large fluctuations) is observed, indicating reliable optimization behavior. Furthermore, compared to standard RLHF pipelines, our PPO variant is more computationally efficient by sharing a frozen LVLM backbone across the policy, value, and reference networks. This design avoids redundant forward passes and reduces both memory usage and computational overhead.}
 
\section{Experiment}

\subsection{Experimental Settings}
\textbf{Datasets:} We consider both a sparse-object dataset (COCO) \cite{lin2014microsoft} and a dense-object dataset (Visual Genome) \cite{krishna2017visual}.  To prepare datasets for LVLM fine-tuning, we randomly augment each image with captioning instructions, as detailed in the Appendix~\ref{instruction}. For COCO, we adopt the $\text{CHAIR}$ metric, which is specifically designed for this dataset and ensures the reliable, consistent evaluation of hallucination. For Visual Genome, we employ $\text{CCEval}$ \cite{zhai2023halle}, an extension of CHAIR that leverages LLM-assisted object extraction to better capture objects beyond the COCO objects, making it well-suited for the Visual Genome benchmark.

For OCS reward computation, the object lists are derived from dataset annotations rather than object detectors. Specifically, for COCO, we use the ground-truth object annotations provided by the dataset. For Visual Genome, we use the ground-truth object annotations together with objects extracted from the reference descriptions following the CCEval protocol \cite{zhai2023halle}. For the detector-based experiments in Section VI, object lists are instead obtained from object detectors to emulate practical deployment scenarios. Specifically, we employ YOLOv8x and YOLOv11x pretrained on COCO, which recognize only the 80 COCO object categories. In real-world deployments, integrating open-vocabulary detectors should be considered to mitigate hallucinations involving non-COCO objects that may otherwise remain undetected.

\textbf{Implementation detail:} We employ InstructBLIP \cite{dai2305instructblip} as our baseline LVLM. InstructBLIP adopts the BLIP-2 architecture \cite{li2023blip2} and is distinguished by its use of Q-former, a Query Transformer designed for instruction-aware training. In this paper, the vision encoder utilized is ViT-g/14 \cite{fang2023eva}, while the LLM of choice is Vicuna-7B \cite{zheng2023judging}. During RL fine-tuning, we initialize the model with the pretrained InstructBLIP checkpoint. Subsequently, we exclusively fine-tune the parameters of our adapters, keeping the image encoder, Q-former, and LLM frozen. 

Our experiments are conducted using the Transformers \cite{wolf-etal-2020-transformers} and PyTorch \cite{paszke2019pytorch} frameworks. For fine-tuning on the dataset, we employ the same tokenizer as InstructBLIP with a vocabulary  $\mathcal{V}$ of the size 32000. Our reward function sets $\alpha$ and $\beta$ to 0.1 and 1, respectively. The soft prompt length is set to 20. The $\gamma$ is set to 0.99. The KL coefficient $\lambda$ is dynamically adjusted, as described in \cite{ziegler2019fine}, with a target KL of 0.05 for stable optimization.    In implementing  PPO, we adopt the default parameters of the Stable Baseline API \cite{stable-baselines3}, with modifications: we gather 4096 transitions and update the PPO loss 5 times for each on-policy step. To enhance training stability, the advantage function used in PPO is approximated using Generalized Advantage Estimation as outlined in \cite{schulman2015high}.  We train the models using the AdamW optimizer with a learning rate of 0.0002 with batch size 64. We leverage 8 Nvidia A6000 GPUs, employing mixed precision and flash attention mechanisms \cite{dao2022flashattention} to enhance training speed. The fine-tuning process typically requires approximately one day to complete. In our experiments, each model is trained using a single run. \rev{The best checkpoint is selected based on validation performance, and this selected model is then evaluated on the test set. All reported results correspond to the performance of this model, averaged over the entire test set.}

\subsection{Experimental Results}

In this section, we present experimental results that highlight five key points: (1) the occurrence of object hallucination and its amplification in detailed captions; (2) the potential of prompt-tuning (demonstrated using hard prompting) to mitigate hallucination; (3) the effectiveness of our framework in reducing hallucination while preserving or even enhancing caption quality when ground truth captions are available, compared to the baseline models and alternative methods; (4) the framework’s robustness when applied to more complex datasets; and (5) Generalization to other LVLMs. 

\begin{table*}[htp]
\centering
\begin{tabular}{@{}c|lcccc@{}}
\toprule
\multicolumn{1}{l|}{\multirow{2}{*}{Type}} & \multirow{2}{*}{Prompt} & \multicolumn{2}{c}{InstructBLIP} & \multicolumn{2}{c}{mPLUG-Owl} \\
\multicolumn{1}{l|}{} & & $\text{CHAIR}_i$(\%)$\downarrow$ & $\text{CHAIR}_s$(\%)$\downarrow$ & $\text{CHAIR}_i$(\%)$\downarrow$ & $\text{CHAIR}_s$(\%)$\downarrow$ \\ \midrule
\multirow{2}{*}{Short} & Generate a short caption of the image. & 2.43 & 3.13 & 22.81 & 60.55 \\
& Create a textual summary for the image. & 4.95 & 6.51 & 22.98 & 61.33 \\ \midrule
\multirow{2}{*}{Long} & Provide a detailed description of the image. & 27.01 & 60.91 & 26.03 & 71.39 \\
& Create a detailed textual summary for the image. & 25.80 & 59.11 & 24.25 & 66.31 \\ \bottomrule
\end{tabular}
\caption{Object Hallucination, gauged by $\text{CHAIR}_s$ and $\text{CHAIR}_i$ metrics, across diverse caption types paired with specific input prompts in the COCO test set. These prompts are designed to elicit both short and long captions. Two distinct methods are illustrated: InstructBLIP and mPLUG-Owl.}
\label{tab:motivation}
\end{table*}

\textbf{The occurrence of object hallucination and its amplification in detailed captions:} We begin by conducting an experiment aimed at demonstrating the presence of object hallucination and its amplification with detailed captions. We design instructions to generate short and long captions using two baseline models: InstructBLIP and mPLUG-Owl. Table \ref{tab:motivation} illustrates the object hallucination measured by CHAIR across various caption types with specific input prompts on the COCO test set. The results indicate that LVLMs experience object hallucination for both short and long captions, with the issue being more pronounced for longer captions. Notably, InstructBLIP exhibits less hallucination with short captions; however, the problem amplifies significantly, around ten times, with longer sentences. Both models show similarly high rates of hallucination in long captions demonstrating the severity of the problem.

\begin{table}[htp]
    \centering
    \begin{tabular}{lccc}
    \toprule
    Method & $\text{CHAIRi} (\%) \downarrow$  &  $\text{CHAIRs} (\%) \downarrow$ & $\text{Coverage} (\%) \uparrow$ \\
    \midrule
    \text{mPLUG-Owl}      &   26.2 & 70.5 & 76.7 \\
    \text{LLaVA}        &   22.5 & 62.7  & \textbf{79.6}\\
    \text{LlaVA-Next} & 13.0 & 32.0  & 68.2 \\
    \text{InstructBLIP}   &   25.8 & 59.1 & 71.3 \\
    \midrule
    \text{Teacher} &   7.5 & 36.4  & NA\\
    \text{CoT}      &   7.8 & 35.7 & NA\\
    \text{Greedy-Decoding}    &  7.8 & 35.5& NA \\
    \text{GPT-Ensemble}     &   13.0 & 51.0& NA \\
    \text{GPT-Teacher}    &   7.8 & 32.0 & NA\\
    \text{LURE}    &   \textbf{5.1} & 21.0 & NA\\
    \midrule
    \text{Hard Prompting}                      &   20.9 & 45.1  & 63.1\\
    \midrule
     \rowcolor{highlight} \text{Ours}                             &   6.8  & \textbf{17.8}  &  70.2\\
    \bottomrule
    \end{tabular}
    \caption{Performance of object hallucination on the COCO dataset. The first row presents the non-fine-tuned LVLM baselines. Starting from the second row, all methods are built upon the InstructBLIP baseline. Specifically, the second row includes hallucination mitigation approaches reported in \cite{zhou2023analyzing}. We additionally report Coverage (i.e., the recall percentage of ground-truth objects mentioned) to quantify recall preservation. The third row presents the results of our Hard Prompting strategy, while the final row reports the performance of the baseline fine-tuned using our RL framework.}
    \label{tab:comparison}
\end{table}

\textbf{Prompt tuning in mitigating hallucination:} We have meticulously curated a series of hard prompts intended to be incorporated at the beginning of input instructions, aimed at minimizing object hallucination in the model's generated captions. Each prompt is meticulously designed to address specific sources of object hallucination, strategically guiding the model away from potential pitfalls. The comprehensive list of prompts is provided in Appendix~\ref{hardprompt}. During the testing phase, we randomly select a single hard prompt to prefix each sample instruction. We conduct captioning using the InstructBLIP baseline model with prefixed instructions. The reported performance metrics reflect the average performance across these instances, focusing particularly on \text{CHAIR} evaluations as shown in Table \ref{tab:comparison} under the label \text{Hard Prompting}.

In comparison to InstructBLIP, we observe that hard prompting can mitigate object hallucination by reducing $\text{CHAIR}_i$ and $\text{CHAIR}_s$from 25.8 to 20.9 (\textcolor{green}{-4.9\%}) and 59.1 to 45.1 (\textcolor{green}{-14\%}) respectively. This suggests that prompt tuning can be used to reduce object hallucination. 

\textbf{Performance of our framework:} Based on the observed effectiveness of hard prompting, we fine-tuned the InstructBLIP model using a learnable soft prompt within our framework to optimize prompt selection on COCO dataset. In Table \ref{tab:comparison}, we present the performance of our proposed method compared to various baselines. The first row represents hallucination of the state-of-the-art LVLM models before fine-tuning: mPLUG-Owl  \cite{li2022mplug}, LLaVA \cite{liu2023visual}, LLaVA-NeXT \cite{liu2024llavanext}, \text{InstructBLIP} \cite{li2023blip}. We show several hallucination reduction approaches on the baseline InstructBLIP in the second rows as presented by \cite{zhou2023analyzing}. The third is Hard Prompting and the last is fine-tuning with our framework.

\begin{table}
    \centering
    \setlength{\tabcolsep}{4pt} 
    \renewcommand{\arraystretch}{1.2} 
    \begin{tabular}{lccc}
    \toprule
    \multirow{2}{*}{Method}  & \multicolumn{3}{c}{Captioning Quality} \\
    \cmidrule(r){2-4}  
    & \text{SPICE} $(\%) \uparrow$ & \text{BLEU} $(\%) \uparrow$ & \text{BERTScore} $(\%) \uparrow$      \\
    \midrule
    \text{mPLUG-Owl}      &  12.5  &  2.7  & 87.4  \\
    \text{LLaVA}         &  13.5  &  3.0  & 87.8  \\
    \text{LLaVA-Next}    &    13.6  &     2.9  &  87.5 \\
    \text{InstructBLIP}    &  10.9  &  1.1  & 85.8  \\
    \midrule
    \text{Hard Prompt}                      &  11.1  &  1.0  & 85.9  \\
    \midrule
    \rowcolor{highlight} \text{Ours}                             &  11.0  &  1.5  & 86.9  \\
     \rowcolor{highlight} \text{Ours+}                             &  \textbf{14.6}  &  \textbf{6.6}  & \textbf{92.4}  \\
    \bottomrule
    \end{tabular}
    \caption{Captioning quality is evaluated using NLP metrics, comparing our approach to other baseline. \textbf{Ours} uses only \text{OCS} and KL divergence, while \textbf{Ours+} incorporates additional metrics: METEOR and BERTScore in the reward function.}\label{tab:captionperformance}
\end{table}

The results demonstrate that the model ,fine-tuned under ours, consistently outperforms all non-fine-tuning baselines across hallucination metrics. Remarkably, our approach enhances $\text{CHAIR}_i$ by \textcolor{green}{+18.9\%} and $\text{CHAIR}_s$ by \textcolor{green}{+41.3\%} compared to the baseline InstructBLIP and notably surpasses the performance of $\text{Hard Prompting}$. Among hallucination reduction approaches, we achieved the top ranking on $\text{CHAIR}_s$ and second place on $\text{CHAIR}_i$, with a very marginal difference compared to the best-performing model, LURE.

Additionally, our method is able to maintain or enhance caption quality across various metrics. Table \ref{tab:captionperformance}  presents our results. The row  \textbf{Ours} demonstrates the use of $\text{OCS}$ and KL divergence, maintaining performance comparable to the base model, InstructBLIP. There is a slight increase in $\text{SPICE}$, $\text{BLUE}$, and $\text{BERTScore}$, which we attribute to the generated captions being more factual, concise, and focused, resulting in shorter and more precise outputs. When ground truth captions are available, incorporating METEOR and BERTScore, as in \textbf{Ours+}, significantly improves caption quality. It is evident that \textbf{Ours+} significantly improves captioning performance across $\text{SPICE}$, $\text{BLEU}$, and $\text{BERTScore}$, surpassing all previous baselines.

\textbf{Extend evaluations to more complex dataset:} We evaluate our model on the Visual Genome dataset using the CCEval metric from Halle-switch \cite{zhai2023halle} to assess its performance in more complex settings with denser object in the image.

\begin{table}[h]
    \centering

    \begin{tabular}{@{}lll@{}}
    \toprule
    Model       & $\text{CCEval}_i  \downarrow$ & $\text{CCEval}_s \downarrow$ \\
    \midrule
    InstructBLIP       &  22.3              & 72.0               \\
    \midrule
     \rowcolor{highlight}  Ours       & \textbf{15.6}       &  \textbf{43.0}      \\
    \bottomrule
    \end{tabular}
    \caption{Performance on the Visual Genome dataset. Note that for Visual Genome, CCEval is selected as the performance metric instead of CHAIR. CCEval provides a more accurate evaluation by extending CHAIR's methodology using GPT-assisted parsing for object extraction, specifically tailored to the Visual Genome dataset.}
    \label{tab:complexDataset}
\end{table}

Table \ref{tab:complexDataset} show performance of our fine-tuned model compare to the baseline InstructBLIP. The result demonstrate that our model substantially mitigates object hallucinations, reducing $\text{CCEval}_s$ from 72.0 to 43.0 (\textcolor{green}{-29.0\%}) and $\text{CCEval}_i$ from 22.3 to 15.6 (\textcolor{green}{-6.7\%}), demonstrating strong robustness beyond the COCO dataset.

\textbf{Generalization to other LVLMs:} To assess the broader applicability, we train another LVLM-LLaVA-under our framework  and evaluate its performance across multiple metrics, as shown in Table \ref{tab:lava}. Our method  significantly reduces object hallucination, reflected in the substantial drop in scores $\text{CHAIR}_i$ (22.5 $\rightarrow$ 7.7) and $\text{CHAIR}_s$ (62.7 $\rightarrow$ 22.2). Additionally, it improves generation quality, BERTScore (87.4 $\rightarrow$ 87.5), and BLEU (3.0 $\rightarrow$ 3.1). These results demonstrate that our approach effectively enhances factual consistency and caption quality, showcasing that our framework generalizes well to other LVLMs.

\begin{table}[h]
\footnotesize
\setlength{\tabcolsep}{2pt} 
\renewcommand{\arraystretch}{1.2} 
\centering
\begin{tabular}{@{}lcccc@{}}
\toprule
\textbf{Model} & $\text{CHAIR}_i \downarrow$ & $\text{CHAIR}_s \downarrow$  & $\text{BERTScore} \uparrow$ & $\text{BLEU} \uparrow$ \\
\midrule
LLaVA          & 22.5              & 62.7                    & 87.4                & 3.0           \\
 \rowcolor{highlight} LLaVA+Ours     & \textbf{7.7}      & \textbf{22.2}          & \textbf{87.5}       & \textbf{3.1}  \\ 
\bottomrule
\end{tabular}
\caption{Comparison of LLaVA baseline and LLaVa fine-tuned using our method.}
\label{tab:lava}
\end{table}

\begin{figure*}[tb]
    \centering
    \includegraphics[width=1.0\linewidth]{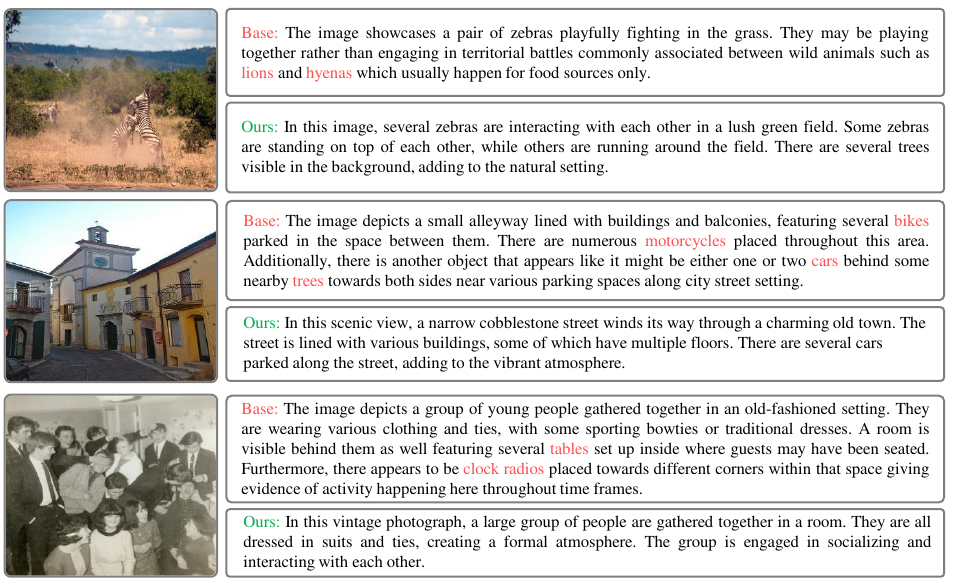}
    \caption{Qualitative comparison between the original InstructBLIP model (\textit{Base}) and the fine-tuned InstructBLIP model trained with the proposed framework (\textit{Ours}) on the COCO dataset. Hallucinated objects are highlighted in \textbf{red}. Compared with the baseline, our method generates captions that are more grounded in the visual content and significantly reduces object hallucinations.}
    \label{fig:qualitativeResult}
\end{figure*}

\textbf{Qualitative Results:}
Figure~\ref{fig:qualitativeResult} presents qualitative examples on the COCO dataset. \textit{Base} denotes the original InstructBLIP model, while \textit{Ours} refers to the fine-tuned InstructBLIP model trained using the proposed framework.

As shown in Figure~\ref{fig:qualitativeResult}, the baseline model frequently introduces objects that are not present in the image, such as, \textit{bikes}, \textit{motorcycles}, \textit{tables}, and \textit{clock}. In contrast, the fine-tuned model generates descriptions that are more faithful to the visual content and avoids mentioning unsupported objects. These examples demonstrate that our framework effectively mitigates object hallucination while preserving the overall descriptiveness and semantic relevance of the generated captions.

\section{Ablation Study}

\textbf{Effectiveness of The Object Consistency Score \text{OCS}:} The OCS plays a crucial role in maintaining high object recall in generated captions. Table~\ref{tabOCS} compares the use of $\text{OCS}$ versus $\text{CHAIR}$ in the reward function. Employing $\text{CHAIR}$ directly leads to a substantial decline in recall (from 71.3 to 20.6), as it overemphasizes precision while neglecting recall. This imbalance allows the LVLM to exploit the reward signal by generating fewer objects to artificially boost precision. In contrast, $\text{OCS}$ mitigates this issue by jointly encouraging both precision and recall, ensuring comprehensive object coverage. The results demonstrate the effectiveness of $\text{OCS}$, achieving 93.2 precision and 70.2 recall, thereby resolving the reward hacking problem.

\begin{table}[htp]
    \centering
    \setlength{\tabcolsep}{2pt} 
    \renewcommand{\arraystretch}{1.2} 
    \begin{tabular}{@{}l|cccc@{}}
\toprule
            Reward  & \multicolumn{1}{l}{\text{Pre} (\%) $\uparrow$} & \multicolumn{1}{l}{\text{Rec} (\%)$\uparrow$} & \multicolumn{1}{l}{$\text{CHAIR}_i$ (\%)$\downarrow$} & \multicolumn{1}{l}{$\text{CHAIR}_s$ (\%)$\downarrow$} \\ \midrule
Base      & 72.9                              & 71.3                           & 27.1                           & 60.9                           \\

CHAIR& 93.7                             & 20.6                           & 6.3                            & 14.4                           \\ 
 \rowcolor{highlight} OCS     & 93.2                              & 70.2                           & 6.8                            & 17.8                           \\
\bottomrule
\end{tabular}
\caption{Comparison of \text{Precision} ($Pre$) and \text{Recall} ($Rec$) between using \text{CHAIR} and \text{OCS} in the reward function.}
\label{tabOCS}
\end{table}

\textbf{Ablation on Incorporating NLP Metrics:} Table \ref{tab:enhanceQuality} illustrates the impact of using different automatic metrics. The baseline model shows high object hallucination with 25.8\% under $\text{CHAIR}_i$. Incorporating the $\text{OCS}$ significantly reduces hallucination down to 6.8\% while maintaining comparable $\text{BERTScore}$ and $\text{BLEU}$ score to the baseline. Adding $\text{BERTScore}$ and $\text{METEOR}$ to the reward function further enhances caption quality, achieving 92.4 in $\text{BERTScore}$ and 6.6 in $\text{BLEU}$ on the COCO test dataset. This ablation study highlights the effectiveness of each component, particularly the $\text{OCS}$'s role in reducing hallucination, and the additional benefits of $\text{BERTScore}$ and $\text{METEOR}$ for optimizing caption quality when reference captions are available.

\begin{table}[h]
    \centering
    \setlength{\tabcolsep}{2pt} 
    \renewcommand{\arraystretch}{1.2} 
    \begin{tabular}{cccc|ccc}
\toprule
Base & OCS & BERTScore & METEOR & $\text{CHAIR}_i$     & $\text{BERTScore}$       & $\text{BLUE}$    \\ \midrule
$\checkmark$                   &                      &                      &                      & 25.8 & 85.8 & 1.1 \\
\rowcolor{highlight} $\checkmark$                     & $\checkmark$                     &                      &                      & \textbf{6.8}  & 86.9 & 1.5 \\
$\checkmark$                     & $\checkmark$                     & $\checkmark$                     &                      & 6.9  & 90.5 & 1.8 \\
\rowcolor{highlight} $\checkmark$                     & $\checkmark$                     & $\checkmark$                     & $\checkmark$                     & 6.9  & \textbf{92.4 }& \textbf{6.6} \\ \bottomrule
\end{tabular}
\caption{The ablation studies examining the impact of OCS, BERTscore and METEOR metrics on the COCO test set}
    \label{tab:enhanceQuality}
\end{table}  
\section{Discussion}




\textbf{On object extraction via detectors:} In our experiments, we utilized visual objects (i.e., objects visible in the image) provided by benchmark datasets to facilitate consistent evaluation. However, when such annotations are unavailable, the effectiveness of hallucination reduction depends on the accuracy of the object detector used—more accurate detections yield stronger mitigation.

To evaluate this effect, we employed YOLOv8x and YOLOv11x to extract objects from COCO images, selecting detections with confidence scores above 0.7 to construct visual object lists. We then fine-tuned InstructBLIP using our framework with YOLO-derived objects and compared the results against models trained with COCO ground-truth objects.

As shown in Table \ref{tab:yolo}, YOLO-based detections substantially reduce hallucinations compared to the baseline (CHAIR\textsubscript{i}: 25.8 → 7.8; CHAIR\textsubscript{s}: 59.1 → 22.3), though they still underperform ground-truth supervision (CHAIR\textsubscript{i}: 6.8; CHAIR\textsubscript{s}: 17.8). Furthermore, YOLOv11x slightly outperforms YOLOv8x in detection mAP (54.7 vs. 53.9) and classification accuracy (79.5\% vs. 79.0\%), yielding lower hallucination scores (CHAIR\textsubscript{s}: 22.3 vs. 25.5). These results underscore that enhancing object detection accuracy directly translates to stronger hallucination mitigation, highlighting the trade-off between automated and ground-truth visual objects.

\begin{table}[h]
\centering
\begin{tabular}{@{}lllll@{}}
\toprule
Model    & Detection mAP & Top1-Acc & $\text{CHAIR}_i$ & $\text{CHAIR}_s$ \\ \midrule
Baseline & -             & -                   & 25.8   & 59.1   \\
YOLOV8   & 53.9          & 79.0                & 7.8    & 25.5   \\
YOLOV11  & 54.7          & 79.5                & 7.8    & 22.3   \\
GT       & -             & -                   & 6.8    & 17.8   \\ \bottomrule
\end{tabular}
\caption{Effect of object detection accuracy on hallucination reduction.
YOLO-based detections greatly reduce hallucination compared to the baseline, with YOLOv11x performing slightly better; ground-truth objects remain most effective}
\label{tab:yolo}
\end{table}

\rev{The quality of object detection directly affects the reliability of the OCS-based reward. Since OCS depends on the overlap between detected visual objects and those mentioned in the generated caption, inaccurate detections (e.g., missed objects or false positives) can lead to biased reward signals. In particular, missing objects may underestimate recall, while incorrect detections may penalize correct captions, thereby weakening the effectiveness of the reward in guiding the policy. Conversely, more accurate detectors provide a more faithful approximation of ground-truth objects, resulting in more reliable precision–recall estimates and stronger supervision. This observation is consistent with our empirical results, where improvements in detector accuracy (e.g., YOLOv11x vs. YOLOv8x) lead to better hallucination mitigation.}

\textbf{Discussion on Utilization and Scalability:} Our study highlights several important insights regarding the utilization and scalability of the proposed framework. First, we observe that automatic feedback provides a viable alternative to human supervision for LVLM fine-tuning, effectively reducing human effort. The performance of our approach currently hinges on automatic metrics such as the OCS; however, as more reliable hallucination detection metrics emerge, they can be readily integrated into the framework with minimal modification. This flexibility underscores the potential for extending our method beyond hallucination mitigation to broader reinforcement learning–based applications in LVLM training.

Another key consideration is computational efficiency. Fine-tuning LVLM inherently requires substantial GPU resources, which can pose practical challenges for large-scale deployment. To mitigate this, our framework was deliberately designed with efficiency in mind—eliminating redundant network duplication and employing parameter-efficient fine-tuning (PEFT). Techniques such as mixed precision training~\cite{micikevicius2017mixed}, efficient attention mechanisms (e.g., xFormers~\cite{xFormers2022}), and distributed training frameworks (e.g., Accelerate~\cite{accelerate}) further enhance scalability and speed.

Despite these optimizations, full fine-tuning still demands considerable computational resources. Interestingly, our findings reveal that LVLMs demonstrate strong prompt-following capabilities. This suggests a promising direction for future research: exploring inference-time adaptation strategies such as prompt engineering to approximate fine-tuning benefits in resource-constrained environments. Larger LVLMs, in particular, are expected to benefit from these strategies, given their inherent prompt understanding and reasoning capacity.

\rev{\textbf{On the caption length:} Our method achieves a clear reduction in hallucinations, with a modest decrease in average caption length (81 vs. 110 tokens). This difference is an expected consequence of our design: by explicitly prioritizing factual consistency, the model suppresses speculative or unverifiable content that often inflates caption length in baseline methods. Importantly, this reduction does not reflect a loss of essential information. Qualitative inspection and metric-based evaluation (e.g., F1, METEOR, BERTScore) confirm that the generated captions remain semantically complete while being more precise and grounded in the visual evidence. In other words, the shorter length primarily results from eliminating redundant or hallucinated details rather than omitting relevant content.}

\rev{\textbf{Potential extension to other LVLM architectures:} Our method is readily extensible to a wide range of LVLM architectures, as it operates entirely at inference time without modifying model parameters or internal structures. It relies only on standard interfaces shared by most LVLMs—namely, the input prompt and token-level outputs (e.g., logits or hidden states). As a result, it can be directly applied to both encoder–decoder and decoder-only multimodal transformers, as long as these outputs are accessible through standard APIs. The approach is independent of the specific visual encoder or language backbone, making it broadly applicable across different model designs. While the extent of improvement may vary with the model’s intrinsic grounding capability, the underlying mechanism remains generally applicable, enabling seamless integration into existing and future LVLMs.}


\textbf{On Inference Runtime:}
\rev{We compare inference latency between hard prompting and soft prompt tuning. The evaluation is conducted on 5,000 COCO test images using a single NVIDIA A6000 GPU. The average wall-time per image is 5.384 seconds for hard prompting and 5.341 seconds for soft prompt tuning. The measured difference (0.043 seconds) is marginal relative to the overall inference time and does not indicate any additional computational overhead introduced by soft prompt approach. This is expected, as soft prompts consist only of a small set of learned embedding tokens prepended to the input sequence, without modifying model depth or requiring additional forward passes.
}

\section{Limitations and Future Directions}
\rev{While our framework effectively mitigates object hallucination, several limitations remain. First, OCS is currently defined over object sets and therefore primarily targets object-level hallucination. Hallucinations involving incorrect attributes (e.g., color) or relations (e.g., spatial or interaction errors) are not explicitly optimized. However, since OCS is fundamentally a precision–recall (F1-based) formulation, it can be extended to structured units such as (object, attribute) pairs or (subject, relation, object) triplets, provided reliable automatic extraction is available. Exploring such extensions is an important direction for future work.

Second, in real-world deployment settings where ground-truth annotations are unavailable, hallucination mitigation depends on the coverage and accuracy of the object detector used to construct visual object sets. Detectors limited to fixed vocabularies (e.g., COCO’s 80 classes) may leave non-COCO hallucinations undetected. Integrating open-vocabulary detection models could further improve robustness in real-world applications.}

Finally, as a minor limitation, the current study does not evaluate variability across multiple random seeds, and the main OCS reward relies on ground-truth object annotations. The detector-based experiments provide an initial assessment of performance under more realistic object extraction, while the current experimental setup enables a controlled evaluation of the proposed reward design. We also note that, although the framework is evaluated across multiple LVLM settings, it does not exhaustively cover all contemporary architectures. Broader multi-seed evaluation, detector robustness, and validation across a wider range of LVLMs are left for future work.
\section{Conclusion}

In conclusion, this paper addresses the persistent challenge of object hallucination in LVLMs for image captioning. Traditional fine-tuning methods, while effective, face scalability issues due to substantial human effort requirements. To overcome this, we propose a novel framework that leverages reinforcement learning (RL) with automatic natural language processing metrics within an MDP framework. This approach minimizes object hallucination while preserving caption quality, achieved through careful architectural design and a tailored reward function. Our framework effectively reduces hallucination compared to the baseline model, InstructBLIP, while maintaining consistent object coverage and caption quality.  The framework offers practical scalability and represents a significant advancement in improving the reliability of LVLMs for image captioning. More broadly, our findings highlight an important direction in multimodal learning—LVLMs can self-correct hallucinations through metric-guided feedback, paving the way for scalable, reinforcement-driven optimization with minimum human intervention.

\section{Acknowledgment}

This work was partly supported by Center for Applied Research in Artificial Intelligence (CARAI) grant funded by DAPA and ADD (UD230017TD) and partly
supported by the National Research Foundation of Korea(NRF) grant funded by the Korea government(MSIT) (RS-2025-24742969, Intelligent Robotic System using Continual Learning and Multimodal Language Model based Multi Attribute
Feedback) and partly supported by Institute of Information \& communications Technology Planning \& Evaluation (IITP) grant funded by the Korean government(MSIT) (No.RS-2021-II211381, Development of Causal AI through Video Understanding and Reinforcement Learning, and Its Applications to Real Environments).

\bibliographystyle{plain}
\bibliography{refs.bib}

\begin{IEEEbiography}[{\includegraphics[width=1in,height=1.25in,clip,keepaspectratio]{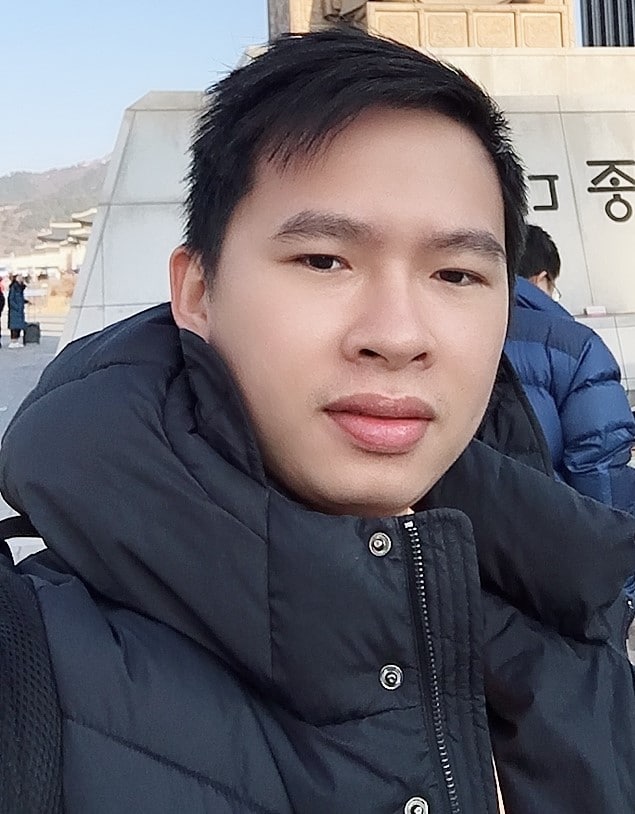}}]{\textbf{Thanh Nguyen}} (Graduate Student Member, IEEE) He received B.Sc degree in electronic and automation engineering from Ho Chi Minh City University of Science and Technology, in 2015. He has been pursuing an M.Sc. and Ph.D. degree at the Korea Advanced Institute of Science and Technology since 2018 under the guidance of Professor Chang D. Yoo. His research interests include machine learning, deep learning, and reinforcement learning.
\end{IEEEbiography}

\begin{IEEEbiography}[{\includegraphics[width=1in,height=1.25in,clip,keepaspectratio]{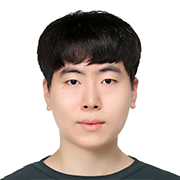}}]{\textbf{Hee Suk Yoon}} (Graduate Student Member) He graduated from the Georgia Institute of Technology with a B.S. in Computer Engineering in 2021. He is currently pursuing a Ph.D. in the field of Artificial Intelligence/Deep Learning at the Korea Advanced Institute of Science and Technology (KAIST) under the guidance of Professor Chang D. Yoo. His research focuses on the reliability of Artificial Intelligence/Deep Learning technologies for various modalities, including images, videos, and natural language. 
\end{IEEEbiography}

\begin{IEEEbiography}[{\includegraphics[width=1in,height=1.25in,clip,keepaspectratio]{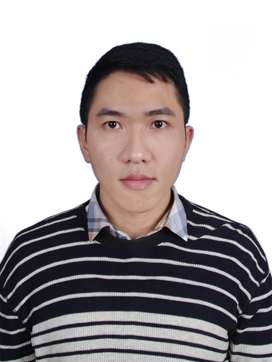}}]{\textbf{Tung M. Luu}} (Graduate Student Member, IEEE) He received B.Sc degree in electronic and telecommunication engineering from Hanoi University of Science and Technology, in 2017, and the M.Sc. degree in electrical engineering from Korea Advanced Institute of Science and Technology, in 2020. He is currently pursuing a Ph.D. degree with Korea Advanced Institute of Science and Technology. His research interests include machine learning, deep learning, and reinforcement learning.
\end{IEEEbiography}

\begin{IEEEbiography}[{\includegraphics[width=1in,height=1.25in,clip,keepaspectratio]{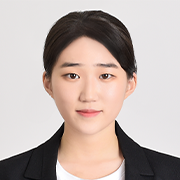}}]{\textbf{Eunseop Yoon}} (Graduate Student Member) She graduated from the Hanyang University, Korea with a B.S. in Electrical Engineering (2021) and from Korea Advanced Institute of Science and Technology (KAIST) with a M.S. (2023). She is currently a Ph.D candidate in KAIST under the guidance of Professor Chang D. Yoo. Her research focuses on enhancing the reliability of AI and Deep Learning tools when working with images, videos, and languages.
\end{IEEEbiography}

\begin{IEEEbiography}[{\includegraphics[width=1in,height=1.25in,clip,keepaspectratio]{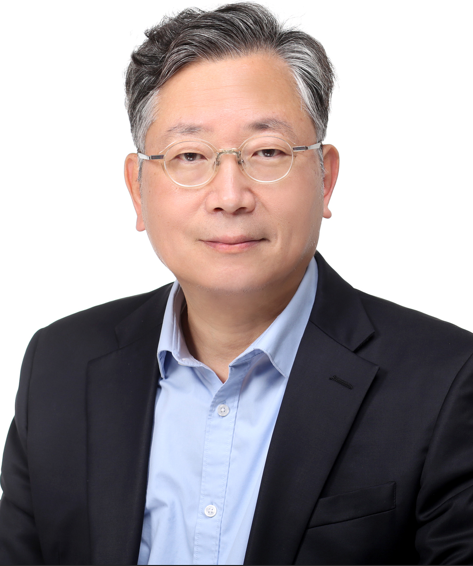}}]{\textbf{Chang D. Yoo}} (Senior Member, IEEE) He received the B.S. from the California Institute of Technology, the M.S. from Cornell University, and the Ph.D. from the Massachusetts Institute of Technology, all in Electrical Engineering. He was a Senior Researcher at Korea Telecom (1997–1999) and has been with KAIST since 1999, where he is a tenured Full Professor in Electrical Engineering and an Adjunct Professor in Computer Science. He has also served as Dean of the Office of Special Projects and the Office of International Relations.
\end{IEEEbiography}

\appendices
\clearpage
\setcounter{page}{1}
\onecolumn

\section{Descriptions of evaluation metrics}

\textbf{BLEU:} BLEU (Bilingual Evaluation Understudy) is a metric employed for assessing the quality of machine-generated translations by comparing them to one or more reference translations. Derived from the concept of precision in n-grams—consecutive sequences of n words—BLEU quantifies the extent to which the generated translation aligns with the reference translations in terms of n-gram overlap \cite{papineni2001method}

\noindent
\textbf{BERTScore:} BERTScore is a technique designed to assess the performance of natural language generation or summarization systems, as introduced by \cite{zhang2019bertscore}. This method gauges the similarity between a reference text and a generated text by leveraging contextualized embeddings derived from BERT (Bidirectional Encoder Representations from Transformers).

\noindent
\textbf{SPICE:} SPICE (Semantic Propositional Image Caption Evaluation) \cite{anderson2016spice} is employed to assess the quality of image captions by evaluating both the semantic content and precision of the generated captions in comparison to reference captions. This metric operates on the hypothesis that semantic propositional content plays a crucial role in human caption evaluation. SPICE introduces an automated caption evaluation method defined over scene graphs, aiming to capture the intricacies of semantic representation in image captions.

\noindent
\textbf{METEOR:} METEOR (Metric for Evaluation of Translation with Explicit ORdering) \cite{banerjee2005meteor} serves as an evaluation metric for machine translation output. This metric calculates the harmonic mean of unigram precision and recall, with recall carrying greater weight than precision. Unlike other metrics, METEOR incorporates additional features such as stemming and synonymy matching, in addition to the standard exact word matching. Its design addresses certain issues identified in the widely used BLEU metric, aiming to improve correlation with human judgment at the sentence or segment level. Notably, METEOR focuses on sentence-level correlation, diverging from BLEU, which seeks correlation at the corpus level.

\section{Description of LVLM models used as baseline}

In this paper, the term Large Vision-Language Models (LVLMs) refers to deep learning models designed to process joint visual and textual data, built upon foundational LLMs. Specifically, LVLMs integrate robust Large Language Models (LLMs) with pre-trained Vision encoders to enhance accuracy in understanding and generating language and vision-related content.

The evaluated LVLMs basically consist of three parts: a visual encoder, an alignment model, and a large language model. All the models have been tuned on collected visual instruction data

\textbf{mPlug-Owl} mPLUG-Owl \cite{ye2023mplug}, is a novel training method that enhances LLMs with multi-modal capabilities by integrating foundational LLM training, a visual knowledge module, and a visual abstractor module. This approach supports various modalities and enhances both unimodal and multimodal abilities through collaborative learning. mPLUG-Owl employs a two-stage training process to align image and text data, leveraging LLM assistance while preserving and enhancing its generative capacities. Initially, the visual knowledge and abstractor modules are trained using a fixed LLM module to align image-text pairs. Subsequently, language-only and multi-modal supervised datasets are utilized to fine-tune a Low-Rank Adaptation (LoRA) module on LLM and the abstractor module while keeping the visual knowledge module frozen.

\textbf{LLaVA} uses a linear projector to map visual token as a soft-prompt into LLM input tokens. LLaVA has a two-stage training, where the initial stage focuses on simple caption pretraining solely for the linear projector, while the subsequent stage finetunes both the projector and LLM on instruction data. Instruction data leverages language-only GPT-4 by inputting visual ground truth from COCO dataset.

\textbf{LLaVA-NeXT} is an enhanced version of LLaVA, featuring improved visual resolution and advanced reasoning capabilities for more sophisticated understanding of both images and videos.

\textbf{InstructBLIP} adopts the BLIP-2 architecture, and is distinguished by its training of a Q-former, which bridges the frozen vision encoder and LLM. InstructBLIP’s instruction fine-tuning spans across 26 distinct datasets.

\section{Instruction Template for detailed image captioning}
\label{instruction}

We use Instruction Templates to generate long, detailed captions. During training, the prompt is randomly selected to query the LVLM. The Instruction Templates are at below:
\begin{itemize}
    \item \textlangle Image\textrangle A detailed image caption: 
    \item \textlangle Image\textrangle A detailed image description: 
    \item \textlangle Image\textrangle Write a long description for the image. 
    \item \textlangle Image\textrangle Describe the content of the image in detail. 
    \item \textlangle Image\textrangle Can you explain clearly what you see in the image? 
    \item \textlangle Image\textrangle Could you describe clearly what you perceive in the photo?
    \item \textlangle Image\textrangle Please provide a detailed depiction of the picture. 
    \item \textlangle Image\textrangle Provide a detailed description of the given image.
\end{itemize}

\section{Hard prompt design} \label{hardprompt}

We have developed a set of "hard prompts" intended to be appended at the beginning of the input instruction, aiming to mitigate object hallucination in the model's generated captions. Each prompt is meticulously crafted to target specific sources of object hallucination, strategically guiding the model away from potential pitfalls. Below is the comprehensive list of prompts:

\begin{itemize}
    \item Directly prohibit object hallucination : "Please don't hallucinate the objects in the image"
    \item Emphasize concrete details: "Provide captions based on specific, easily identifiable elements in the image."
    \item Prioritize realism: "Generate captions that reflect plausible scenarios and avoid fantastical or improbable elements."
    \item Stick to visible entities: "Describe only what is clearly visible in the image and avoid making assumptions about hidden or obscured objects."
    \item Be conservative in interpretation: "Refrain from extrapolating beyond what is evident in the image; captions should stay closely tied to observable elements."
    \item Avoid creative interpretations: "Discourage the generation of captions that involve imaginative or metaphorical representations of the scene."
    \item Limit descriptive scope: "Keep captions focused on the central objects or subjects in the image, avoiding unnecessary details or peripheral elements."
    \item Minimize speculative language: "Generate captions with certainty, avoiding speculative language or uncertain descriptions of the depicted scene."
    \item Resist contextual speculation: "Do not create captions that rely on external context or background information not present in the image."
    \item Steer clear of abstract concepts: "Refrain from incorporating abstract or conceptual ideas into the captions; stick to tangible, visible elements."
    \item Encourage literal language: "Favor literal and straightforward language in captions, avoiding figurative expressions or interpretations."
\end{itemize}

\section{Motivation of using Reinforcement Learning from automatic metrics} 
Our motivation for employing RL stems from the need to minimize human effort while ensuring effectively reduct hallucination. 

Traditional approaches to mitigating hallucinations often require identifying specific sources of hallucination and designing targeted strategies to counter them. While effective, these methods are labor-intensive. Data-driven alternatives like supervised learning provide some level of automation but rely heavily on labeled datasets, which still require significant human input for data annotation and curation—an increasingly costly and time-intensive process, particularly for large-scale models.

In contrast, reinforcement learning in the literature not only demonstrates strong alignment capabilities for LVLMs in tasks like image captioning but also offers a promising path to automation by significantly reducing the need for explicit labels (e.g., relying only on simple binary feedback for reward modeling). We are motivated to push this approach to its limits by completely eliminating human-labeled data, fully leveraging RL’s potential through the exclusive use of automatic metrics to reduce hallucinations. These metrics are gradually improving in their alignment with human feedback in terms of both accuracy and reliability. Our approach enables the model to iteratively refine its outputs based solely on automatic feedback, providing an efficient and scalable solution that aligns with the trend toward larger LVLMs.

\section{Training settings for prompt fine-tuning}

We present parameter settings in our framework during the training phase. Here we use 8 A6000 50G GPU for training. The information is shown in table below:

\begin{table}[htp]
    \centering
    \begin{tabular}{lc}
    \toprule
    Parameters &     \\
    \midrule
    \text{Do sample (greedy decoding)} & False \\
     \text{Num beams} & 1 \\
    \text{Max Soft Prompt Length $n$}      &   20 \\
    \text{Max Episode Length $H$}        &   150  \\
    \text{Mini-Batch Size}   &    64  \\
    \text{PPO Learning Rate} &   0.0002  \\
    \text{Kullback–Leibler (KL) Initial Coefficient $\lambda$}      &   0.1  \\
    \text{KL Target Coefficient}    &  0.05   \\
    \text{Number of Training Epochs}     &   50   \\
    \text{Reward Weight for Meteor $\alpha$}    &   0.1   \\
    \text{Reward Weight for BERTScore $\beta$}    &   1   \\
    \text{Vocabulary Size $\mathcal{V}$}                      &   32000   \\

    \bottomrule
\end{tabular}
    \caption{Training parameters of our framework for the COCO dataset}
\end{table}

\section{Quanlitative Result}

We present more quantitative results on the COCO dataset. The \textbf{Base} refers to the InstructBlip model, while \textbf{Our} is the fine-tuned version of InstructBlip using our framework. Hallucinated objects are highlighted in bold and red for clarity.

\begin{figure}[H]
    \centering
    \includegraphics[width=1\linewidth]{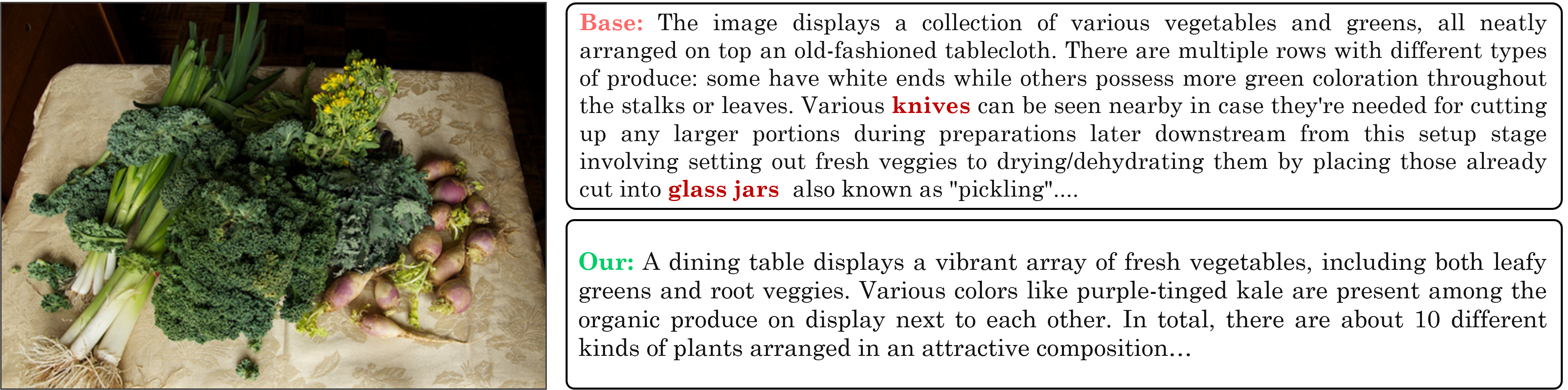}
    \includegraphics[width=1\linewidth]{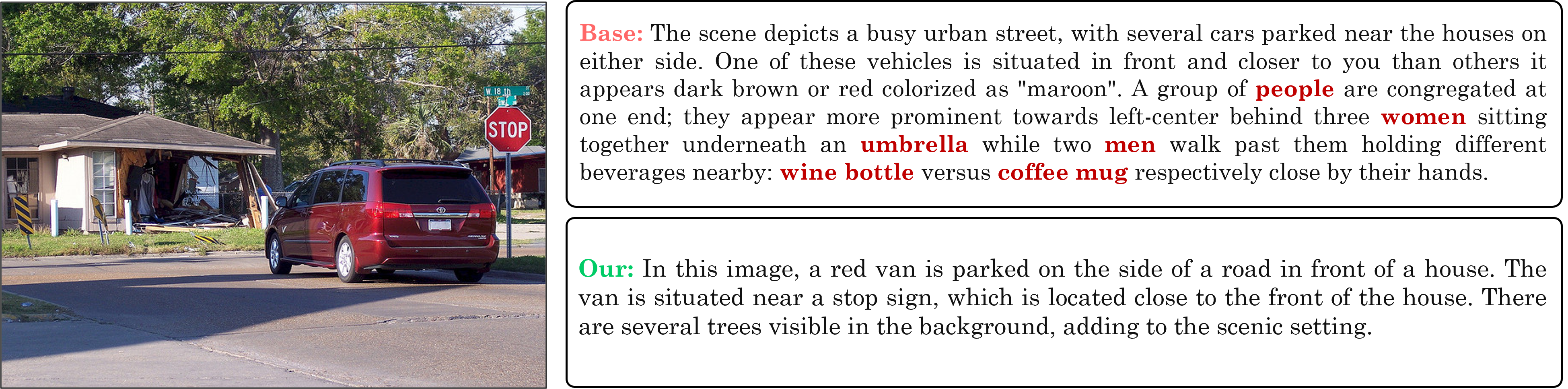}
    \includegraphics[width=1\linewidth]{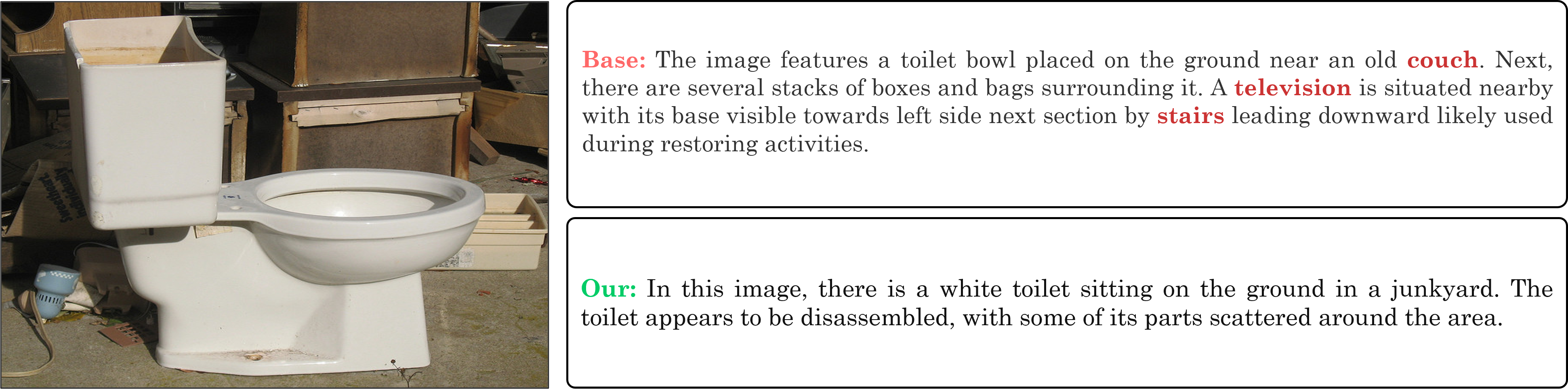}
    \includegraphics[width=1\linewidth]{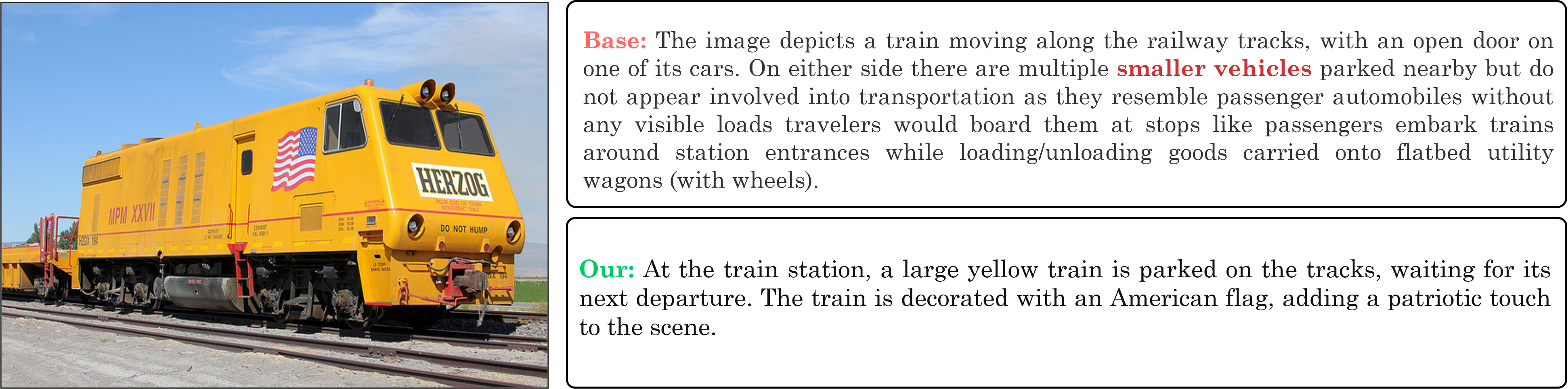}
\end{figure}

\begin{figure}[tb]
    \centering
    \includegraphics[width=1\linewidth]{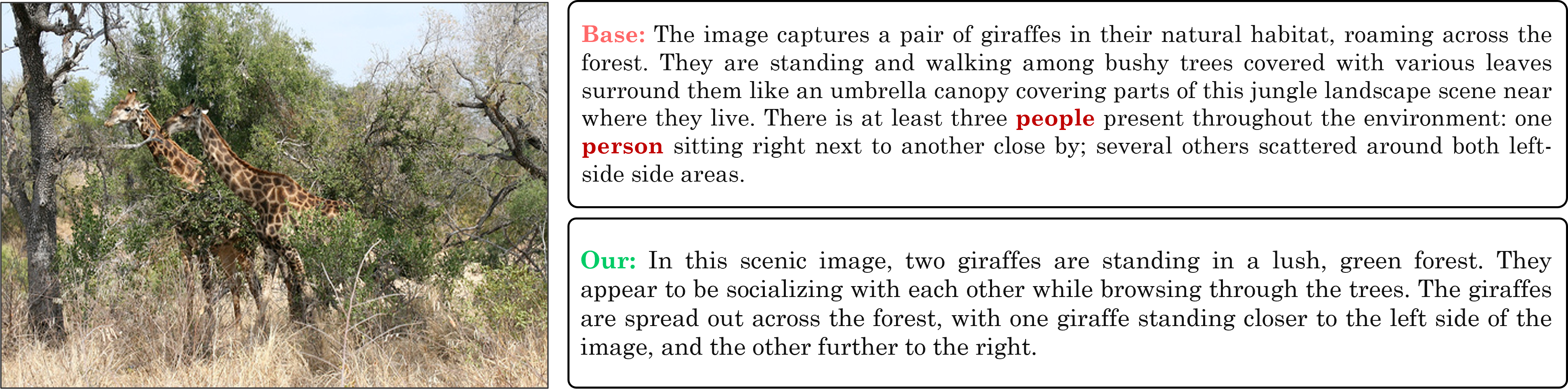}
    \includegraphics[width=1\linewidth]{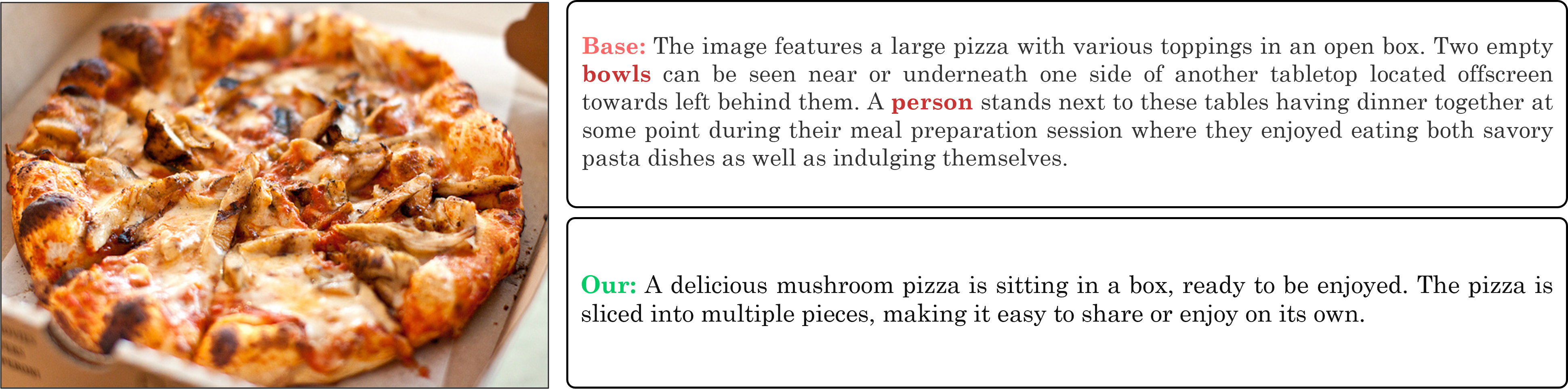}
    \includegraphics[width=1\linewidth]{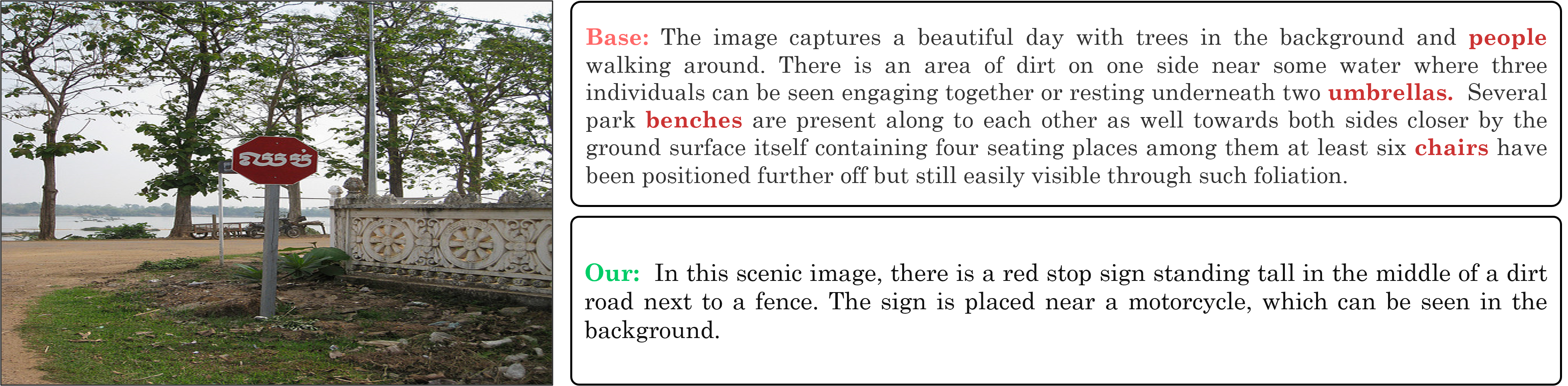}
    \includegraphics[width=1\linewidth]{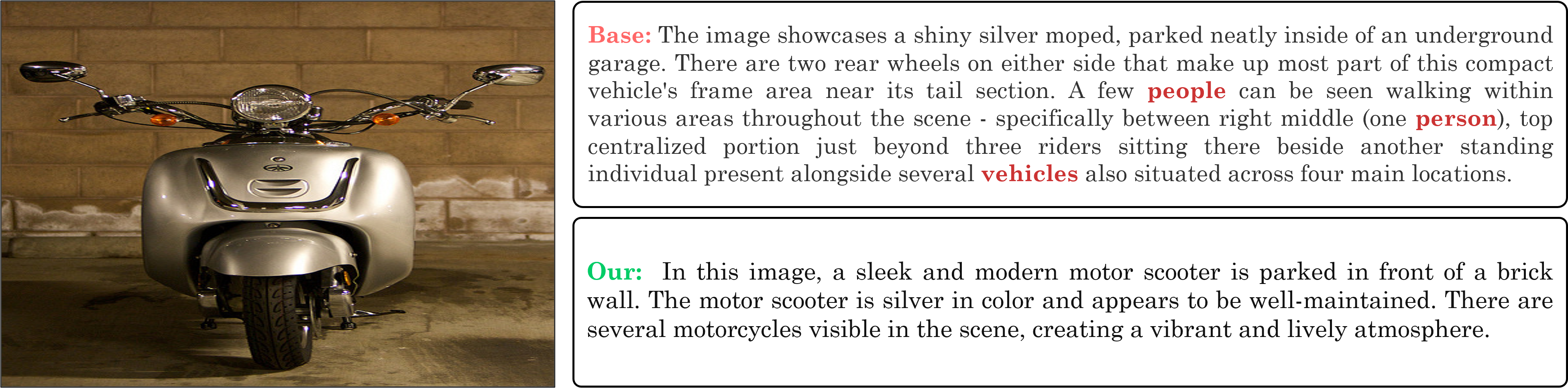}
\end{figure}

\EOD
\end{document}